# Learning Representations of Satellite Images with Evaluations on Synoptic Weather Events

**Ting-Shuo Yo[1,4], Shih-Hao Su[2], Chien-Ming Wu[1], Wei-Ting Chen[1], Jung-Lien Chu[3], Chiao-Wei Chang[1], and Hung-Chi Kuo[1]**

[1]National Taiwan University.

[2]Chinese Culture University.

[3]National Science and Technology Center for Disaster Reduction.

[4]DataQualia Lab. Co. Ltd.

**Key Points:**

- The features of satellite images learned by the convolutional autoencoder performed the best in multiple weather classification tasks.

- The PCA is a powerful feature learner for high hit rates, but it came with higher false alarms and didn't benefit from high-resolution data.

- The proposed framework combined representation learning algorithms with explainable classification methods and can be applied to more complicated problems.

**Abstract**

This study applied representation learning algorithms to satellite images and evaluated the learned latent spaces with classifications of various weather events. The algorithms investigated include the classical linear transformation, i.e., principal component analysis (PCA), state-of-the-art deep learning method, i.e., convolutional autoencoder (CAE), and a residual network pre-trained with large image datasets (PT). The experiment results indicated that the latent space learned by CAE consistently showed higher threat scores for all classification tasks. The classifications with PCA yielded high hit rates but also high false-alarm rates. In addition, the PT performed exceptionally well at recognizing tropical cyclones but was inferior in other tasks.

Further experiments suggested that representations learned from higher-resolution datasets are superior in all classification tasks for deep-learning algorithms, i.e., CAE and PT. We also found that smaller latent space sizes had minor impact on the classification task's hit rate. Still, a latent space dimension smaller than 128 caused a significantly higher false alarm rate.

Though the CAE can learn latent spaces effectively and efficiently, the interpretation of the learned representation lacks direct connections to physical attributions. Therefore, developing a physics-informed version of CAE can be a promising outlook for the current work.

**Plain Language Summary**

Our work compared classical and AI-based methods of deriving features from satellite images. We used the learned features to identify a few weather events that are defined in very different ways. The results showed that the AI-based methods, especially CAE, performed the best among most tasks and can be improved using higher-resolution images. The classical method, PCA, had a similar performance in "identifying an event when it actually happened" but suffered from more false alarms. Finally, we look to improving the CAE with better interpretability in terms of physics in the future.

## 1 Introduction

Satellite imagery is an essential tool for weather diagnosis and forecasting. It enables meteorologists to overview the large-scale and synoptic weather systems and their movement. In addition, the imagery allows the monitoring and detection of smaller-scale phenomena such as convective cells, thunderstorms, and fog. As the resolution and coverage of satellite imagery increased over time, the amount of data also grew significantly. Besides allocating more computational resources for processing satellite data, we can also leverage algorithms to derive features from a large amount of data.

Representation learning is a machine learning subfield focusing on learning the internal structure of data to extract useful information when building classifiers or other predictors (Bengio, 2013). It is a process of learning a parametric mapping from the original input data domain to a space of feature vectors. The early purpose of representation learning, or feature extraction, was to reduce the data dimension to a manageable size. Zhong et al. (2016) reviewed the development of data representation learning and categorized the approaches with four dimensions: linear or nonlinear, supervised or unsupervised, generative or discriminative, and global or local. The linearity of algorithms indicates the relationship between the original data space and the latent space of the representations. The supervised and unsupervised methods are distinguished by whether they require the data to have pre-specified labels. Discriminative approaches draw boundaries in the data space, while generative ones try to model how data is placed throughout the domain. The difference between global and local algorithms is that the former attempts to preserve geometry at all scales while the latter aims for local geometry (Silva & Tenenbaum, 2002).

The earliest representation learning method can be dated back to Pearson's principal component analysis (PCA) in 1901 (Pearson, 1901), a linear, unsupervised, generative, and global feature learning method. Later, in 1936, Fisher proposed linear discriminant analysis (LDA) as a linear, supervised, discriminative, and global representation learning algorithm (Fisher, 1936). Afterward, several approaches were proposed to establish nonlinear mappings between the original data and the learned features. For example, kernel PCA (Schölkopf et al., 1998) and generalized discriminant analysis (GDA. Baudat & Anour, 2000) are the nonlinear counterparts of the PCA and LDA. The emphasis on preserving local geometry in the feature vector space rose with the studies of manifold learning (Roweis & Saul, 2000; Tenenbaum et al., 2000). The manifold learning algorithms try to solve an eigenproblem for embedding high-dimensional points into a lower-dimensional space by defining the distance between points in various ingredients (Bengio et al., 2003; Izenman, 2012). In addition to manifold learning, artificial neural networks (ANN) were employed for nonlinear dimension reduction (Kohonen, 1990; Hinton & Salakhutdinov, 2006). Later, Hinton et al. (2006) introduced the concept of "deep learning," which researchers (Zhong et al., 2016; Khastavaneh & Ebrahimpour-Komleh, 2020) suggested a division between "shallow" and "deep" models. The emergence of deep neural networks was considered to make the learned representations go beyond the data's features and form abstract characteristics (Bengio, 2009).

Along with the development of dimension reduction techniques, the focus of algorithms has shifted in a two aspects. First, the interest in the related methods changed from dimension reduction to finding the internal manifolds of the data. Second, the purpose of the feature vectors shifted from task-specific characteristics to abstract, task-invariant representations. Moreover, researchers have developed core principles of good representations, i.e., *distributed*, *disentangled*, and *abstract and invariant* (Bengio et al., 2013; Le-Khac et al., 2020). The distributed property means good representations should represent as much information as possible while keeping the latent space small. The disentangled property requires an algorithm to form a latent space that can capture as many factors and discard as little data as possible. Finally, the property of abstrction and invariant prefers the representations to be generalizable and robust to small and local changes. In additional to the deriable properties, Bengio et al. (2013) discussed the criteria for evaluating learned representations and pointed out that deep learning approaches have succeeded in multi-task learning and domain adaptation (Krizhevsky et al., 2012; Collobert et al., 2011). This concept inspired us to apply representation learning algorithms to satellite observations and to evaluate the learned features against various atmospheric phenomena.

The rapid development of remote sensing technology has increased the availability of large-scale satellite datasets. With machine learning gaining more and more attention in scientific research, several attempts have been made to apply deep learning to satellite data. For example, object recognition in satellite images is essential for geographical information retrieval and leads to land management and ecology applications (Lu et al., 2017; Jean et al., 2019; Proll, 2019; Alshahrani et al., 2021; Valero et al., 2021). Researchers also applied deep learning algorithms to satellite images to detect tropical cyclones (Pradhan et al., 2017; Chen et al., 2019; Zheng et al., 2019), atmospheric rivers (Chapman et al., 2019), horizontal visibility (Amiri and Soleimani, 2022), and air quality (Sorek-Hame et al., 2022). Despite these efforts, few attempts have been made to explore the representations learned with deep neural networks. In this study, we apply representation learning algorithms to satellite images and evaluate the learned features by classifying multiple atmospheric phenomena. In the designed experiments, we investigated the Convolutional Autoencoder (CAE) and pre-trained Residual Networks (ResNet) and compared the results to the classical PCA.

The representation learning methods, the datasets, and the experimental design are described in the following section. The evaluations of the classification of multiple weather events are summarized in the Results section, followed by discussions and concluding remarks.

## 2 Methods

In this study, we investigated three practices of representation learning, namely Principal Component Analysis (PCA), Auto-Encoder with convolutional kernels (AE), and pre-trained Residual Network (PT). The following subsections introduce each approach and why we choose it for our task.

### 2.1 Principal Component Analysis

Since its first introduction by Karl Pearson in 1901, Principal Component Analysis (PCA) has been widely used as a pattern discovery tool in various scientific fields.

Theoretically, PCA can be thought of as fitting a p-dimensional ellipsoid to the data, where each ellipsoid axis represents a principal component. The fitting process can be mathematically achieved by performing eigendecomposition on the covariance matrix.

Though Pearson's work was the first documented, scientists in the early 20th century came up with similar ideas with different names. For example, researchers use the term empirical orthogonal function (EOF) for the same method in meteorology and geophysics. This approach was widely applied to climate research and resulted in significant findings such as ENSO (Trenberth, 1997).

There have been several improvements in PCA in the past 100 years, and we want to address a few milestones that led to the PCA implementation used in our work. The first improvement for numerical PCA is using the singular value decomposition (SVD) to replace the eigendecomposition. The SVD is a factorization method that generalizes from a square-normal matrix to any *n x m* matrix. The SVD provides a stable numerical solver for matrix factorization, but the computational cost is still considerable when the data dimension is high. For example, the data size can be too large to be stored locally and computed simultaneously. Ross and colleagues introduced an incremental learning approach that enables us to apply PCA to such datasets (Ross et al., 2008). In other cases where the data dimension is too high to be factorized efficiently, the Randomized SVD, a low-rank matrix approximation algorithm introduced by Halko and colleagues, vastly increases the computational efficiency (Halko et al., 2011).

This study used incremental PCA with a randomized SVD solver implemented in the scikit-learn package (Pedregosa et al., 2011). Thus, we managed to project the 30 years of satellite images (256 x 256 pixels) into vectors with the desired length.

## 2.2 Autoencoder

The autoencoder (AE) is an artificial neural network (ANN) used to find a latent space that can represent the data efficiently. An autoencoder consists of two parts: an encoder that projects the original data into the latent space and a decoder that projects vectors from the latent space into the original dimension. The two sub-networks are then trained together with adequately designed objective functions to preserve specific properties in the latent space. For example, such an autoencoder can serve as an efficient compression model for similar data by minimizing the root-mean-squared error (RMSE) between the original data and the model output. The vectors in the latent space can be seen as abstract representations of the original data. The flexibility of ANNs allows users to learn a latent space with desired properties by choosing the corresponding loss function and ANN architecture.

Integrating the convolutional kernels in ANN is one of the breakthroughs in image recognition (LeCun et al., 1989). In image processing, the kernel, also known as the convolution filter, is a small matrix that operates on original image elements and creates a new image. Such a process is a form of mathematical convolution referred to as image convolution.

This study used the convolutional autoencoder (CAE) with the objective function of minimizing RMSE to encode the satellite images into a latent space. The algorithm design and sample code can be found in our FAIR Data Compliance (Yo, 2023).

## 2.3 Pre-trained model

Pre-training neural network models with large datasets is a critical technique in convolutional neural network research (Krizhevsky et al., 2012). This approach arose from the discovery that the learned features on one computer vision task can be transferred to another and led to the studies of general visual representation learning (Kolesnikov et al., 2020). Though He et al. (2018) demonstrated that pre-trained models did not perform better than those trained from scratch, Hendrycks et al. (2019) have shown that pre-training can improve model robustness and uncertainty. Despite the debates, fine-tuning models pre-trained with large datasets is common in computer vision and natural language processing (Han et al., 2021).

In this study, we used a 50-layered residual network (ResNet50) pre-trained with ImageNet (He et al., 2016) and BigEarthNet (Neumann et al., 2019; Sumbul et al., 2019), which can take images of any size and map them into feature vectors in the length of 2,048.

## 3 Data and Experiment Design

This study used the Gridded Satellite dataset (GridSat-B1, Knapp et al., 2011) for representation learning. And since we used the synoptic weather events to evaluate the effectiveness of the learned representations, an open data set of atmospheric events near

Taiwan (TAD, Su et al., 2022) was used as the source of information. A brief introduction of the data and preprocessing procedures is discussed in the following sections.

## 3.1 GridSat-B1 CDR

Gridded Satellite data used in our study are gridded International Satellite Cloud Climatology Project (ISCCP) B1 data on a 0.07-degree latitude equal-angle grid. Satellites are merged by selecting the nadir-most observations for each grid point. The Geostationary IR Channel Brightness Temperature (BT)- GridSat-B1 Climate Data Record (CDR) provides global BT data from geostationary Infrared (IR) satellites.

## 3.2 Weather Events

The Taiwan Atmospheric Event Database (TAD, Su et al., 2022) contains everyday synoptic weather events over the Taiwan area from 1980 to 2020. We selected four types of events in TAD, i.e., front, tropical cyclones, north-easterlies, and south-westerlies. A brief introduction of these events and their definitions in TAD is described as follows.

## 3.2.1 Front (FT)

Weather fronts represent the transition zone between two air masses. Across a front, there can be significant variations in temperature and wind direction. Although the fronts were heavily studied and a few methods existed to define the front objectively, they may not be suitable for the subtropical fronts in Taiwan due to the differences in the thermodynamic properties (Chang et al., 2019). Therefore, Su and colleagues defined a rectangle covering the Taiwan and nearby areas (21° to 26°N, 119° to 123° E). Based on the daily surface map issued by the Central Weather Bureau (CWB) at 00Z (8:00 LTC), the front event is defined whenever the labeled front system on the surface map passes through this rectangle.

### 3.2.2 North-easterlies (NE)

The north-easterlies in the Taiwan area are part of the winter monsoon in East Asia and influence the precipitation in Taiwan's northern part during winter. In TAD, Su et al. used the daily average wind of the Pengjiayu weather station as the indicator of the north-easterlies. The day is labeled an NE event if the average wind direction is between 15 to 75 degrees and the wind speed is above 4m/s.

### 3.2.3 South-westerlies (SWF)

Like the north-easterlies, the south-westerlies in Taiwan represent the large-scale circulation pattern in summer. The TAD used the reanalysis wind field at 850hPa provided by the Nation Centers for Environmental Prediction (NCEP) as a reference due to the lack of weather stations in the upstream region. Su et al. derived the averaged wind properties in a rectangular area between 16° to 22.5° N and 110° to 120°E and labeled an averaged north-eastward wind with wind speed greater than 3m/s as an SWF event.

### 3.2.4 Heavy Rainfall (HR)

The heavy rainfall events are defined by precipitation records of the CWB weather stations. We labeled an HR event while any weather station recorded more than 10mm/hr precipitation within a day. This definition differs from CWB's official operation. Specifically, we lowered the threshold from 99% percentile rank to 90% to create a balanced event record.

3.2.5 Tropical Cyclones in the Northwestern Pacific Ocean (NWPTC)

The International Best Track Archive defines the typhoon events in TAD for Climate Stewardship (IBTrACS, Knapp, et al., 2010) from the World Data Center for Meteorology (WDC). Su et al. categorized the typhoon events as within 100km, 200km, 300km, 500km, and 1000km of Taiwan's coastline. The TAD also defined an event that there existed tropical cyclones over the Northwestern Pacific Ocean, NWPTC, as the IBTrACS records being within the range of 0° to 60°N, 100° to 160° E. We choose the NWPTC as one classification task for learned representations.

As explained above, the five chosen weather events are defined in various ways. For example, though FT and NWPTC are specified manually by human experts, the IBTrACS used for NWPTC is a visible point, while the weather front is an imaginary line. Moreover, NE and SWF are wind-field-based events, but they are defined by one weather station and a large region. Finally, the heavy-rainfall events are depicted with multiple weather stations. These five events are selected to represent the common weather types in the Taiwan area and different ways of definitions.

We selected GridSat-B1 data and the weather events during 2013 ~ 2016 for further analysis, and Table 1 summarizes the counts and frequency of the five events.

Table 1. The counts and frequency of the selected events during 2013~2016.

| Event | Counts | Frequency |
|-------|--------|-----------|
| FT | 244 | 0.17 |
| NE | 471 | 0.32 |
| SWF | 406 | 0.28 |
| HR | 520 | 0.36 |
| NWPTC | 702 | 0.48 |

## 3.3 Experiment Design

The complete experiment design of this study is shown in Figure 1. In the preprocessing step, the original GridSat-B1 dataset was cropped to 0 - 60 N and 100 - 160 E and then rescaled to float numbers between 0 and 1 (divided by 255). Afterward, we used the bilinear interpolation algorithm to interpolate the original resolution from 864x864 to 256x256 and 512x512.

In the representation learning step, we applied PCA, CAE, and the pre-trained RestNet50 to the preprocessed data. Each method resulted in a set of feature vectors of length 2048. Finally, we use the feature vectors as the independent variables and a simple linear classifier, the logistic regression, to identify the five weather events described above.

The logistic regression is a statistical model that models the probability of an event. Like linear regression, logistic regression formulates the linear combination of independent variables and outputs a prediction. Unlike linear regression, the logistic regression model uses the linear formulation's logit function to model the dependent variable's log odds. Hence, the logistic regression prediction indicates the probability of the dependent variable and can be used to perform binary classification (Hastie et al., 2009).

The classification process was evaluated with the 10-fold cross-validation scheme (Hastie et al., 2009). Furthermore, we focused on three metrics commonly used in forecasting, i.e., the hit rate, false-alarm rate, and the threat score (Jolliffe and David, 2011). The workflow of the experiment is illustrated in Figure 1.

We designed a series of experiments with the same workflow. Experiment 1 is based on preprocessed GridSat-B1 dataset with a resolution of 256x256 and serves as the baseline. Experiment 2 is similar to experiment 1 but with a data resolution of 512x512 to examine the performance of algorithms under better data resolution. In experiment 3, we conducted the same workflow with varied feature vector sizes, ranging from $2^{12}$(2048) to $2^2$(4). Because the pre-trained models (PT) have a fixed feature vector size, they are not included in experiment 3. The results of the experiments are shown in the following section.

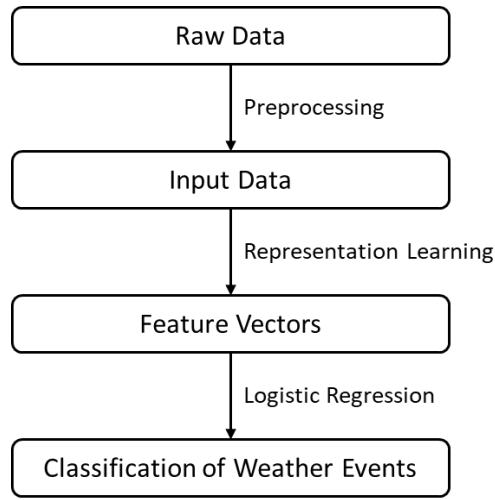

Figure 1. The flow chart of the experiment design.

## 4 Results

### 4.1 Experiment 1: the baseline

Figure 2 summarizes the evaluation metrics of experiment 1, and a full table can be found in the supporting information. The probability of prediction (POD; also known as the hit rate), false-alarm rates (FAR), bias, and critical success index (CSI; also known as the threat score) are shown in the performance diagram (Roebber, 2009). The weather events are represented by different symbols and algorithms by colors. Figure 2 shows that features derived from PCA give a slightly higher hit rate (POD), and those from CAE yield the lowest false alarm (high success ratio). Regarding the threat score, the green symbols appear more upper-right than other colors, which indicates that CAE outperforms other methods in all weather events. The results suggest that the deep neural network models with convolutional kernels can learn proper representations for multiple classification tasks.

Moreover, they yield better performance and show consistent advantages across different weather events.

When looking at the classification metrics between weather events, we found the representations learned from the satellite images did best at identifying SWF, NE, and tropical cyclones. The HR events were less relevant to the satellite features, which did worst on the FT events. Such results were consistent with the domain knowledge. The SWF events are sophisticatedly defined and usually associated with a bright cloud band within a specific region. The NWPTC events also have solid visual characteristics in the satellite, though their locations may vary case by case. As for the HR events, which were supposed to be associated with the visible cloud, the cloud pattern at 00Z might not represent convective clouds developed later and hence caused misclassifications. We further tested the same experiments with the satellite images of 12Z, and the results are similar (the full table can be found in the supporting information). Though the NE events were defined similarly to the SWF, the cloud patterns with the winter monsoon were usually of lower altitudes and thus not as significant as SWF to the infrared sensors. And finally, since the definitions of the FT events were subjective and didn't always associate with the cloud, we are not surprised that the feature vectors learned from satellite images cannot detect it accurately enough.

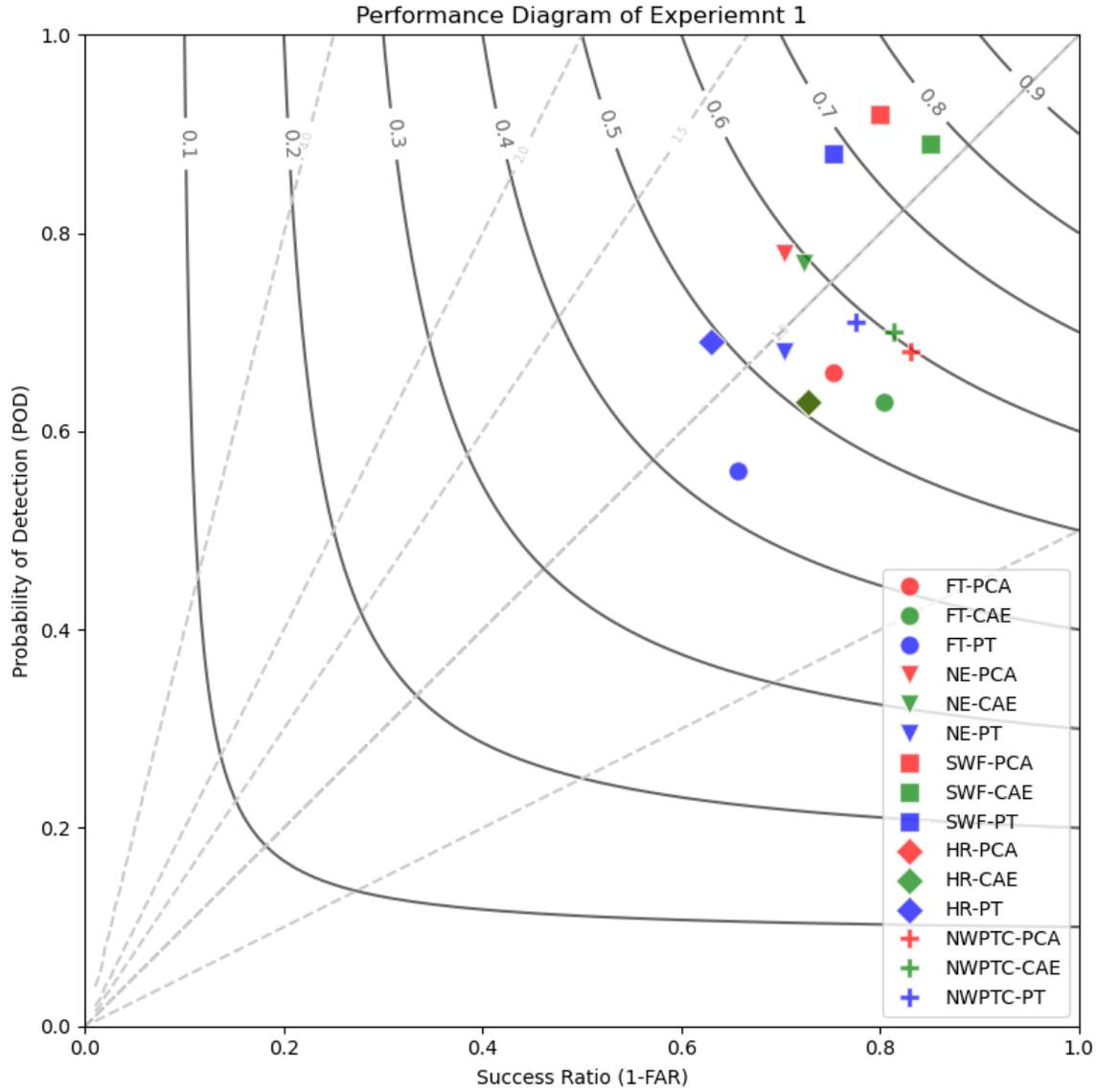

Figure 2. The performance diagram (Roebber plot) of experiment 1. The standard metrics of binary classification tasks, i.e., probability of detection (POD; also known as the hit rate), false alarm ratio (FAR) or its opposite, the success ratio (SR), bias and critical success index (CSI; also known as the threat score) are represented as the x-axis, y-axis,

the solid contours, and the dashed lines. The weather events are shown as different symbols, while algorithms are shown in different colors.

4.2 Experiment 2: the resolution of the satellite images

Main experiment 1 is conducted on the dataset of 256x256 resolution. While there will be more and more high-resolution satellite images available as time goes by, we wanted to check whether the learning algorithms can perform better using high-resolution data. Thus, we conducted the same set of tests on the dataset of 512x512 resolution.

The Roebber plot of experiment 2 is shown in figure 3. The relative performance for the high-resolution experiment is similar to experiment 1, where CAE also gave the highest threat scores except for the NWPTC events. Figure 4 shows the evaluation metrics of the classification with the high-resolution data subtracted by the corresponding values of the low-resolution configuration. As indicated in Figure 4, the dataset with a higher resolution overall has a better performance than experiment 1 for CAE and PT. However, PCA didn't seem to benefit from the higher-resolution dataset, while It is commonly believed that higher-resolution satellite images could provide more details about the atmospheric phenomenon. These results implied that CAE could be a better choice for researchers who wants to take advantage of the increasing availability of high-resolution datasets. The full table of the results of the high-resolution experiment can be found in the supporting information.

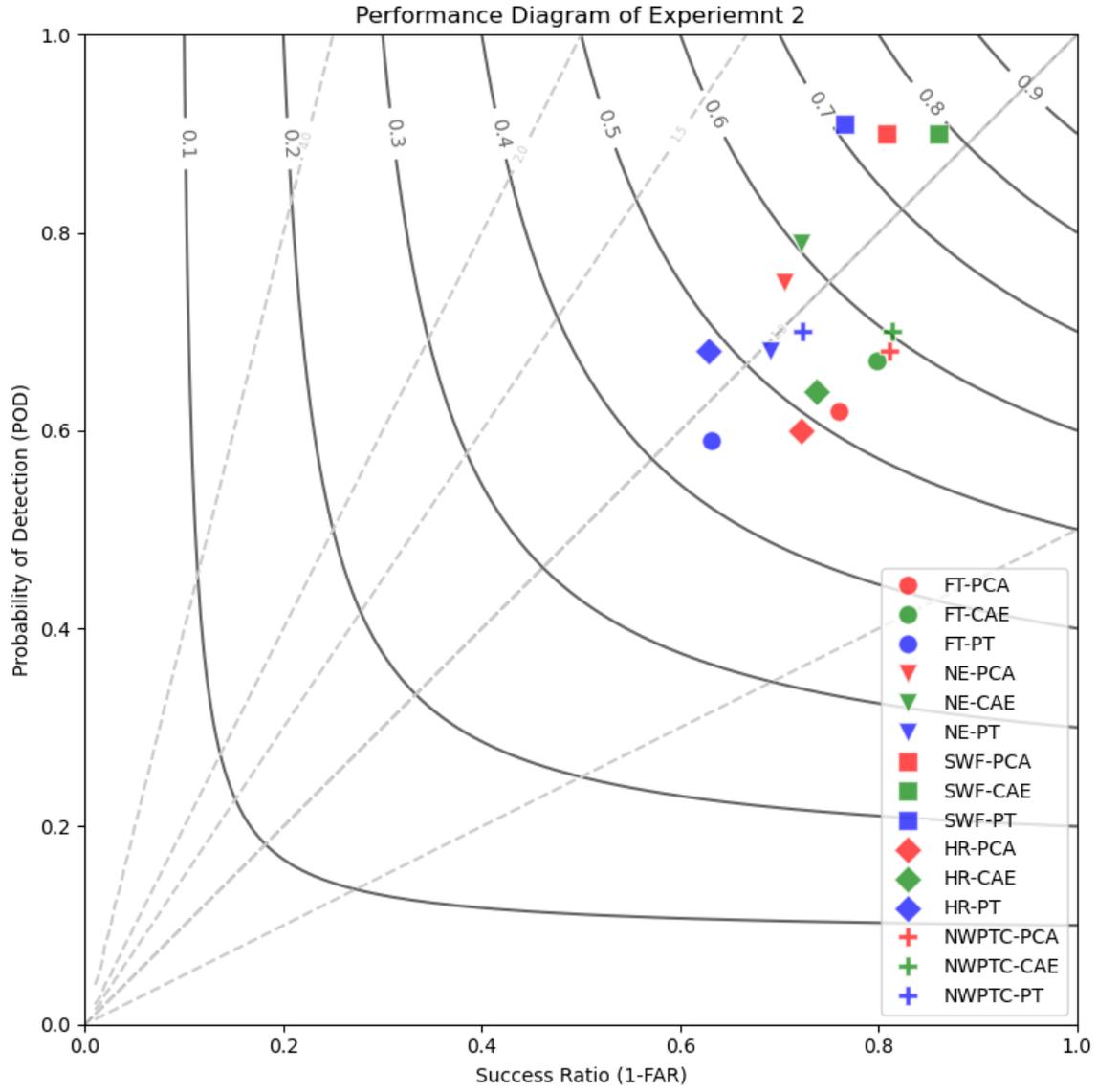

Figure 3. The performance diagram (Roebber plot) of experiment 2. The weather events

are shown as different symbols, while algorithms are shown in different colors.

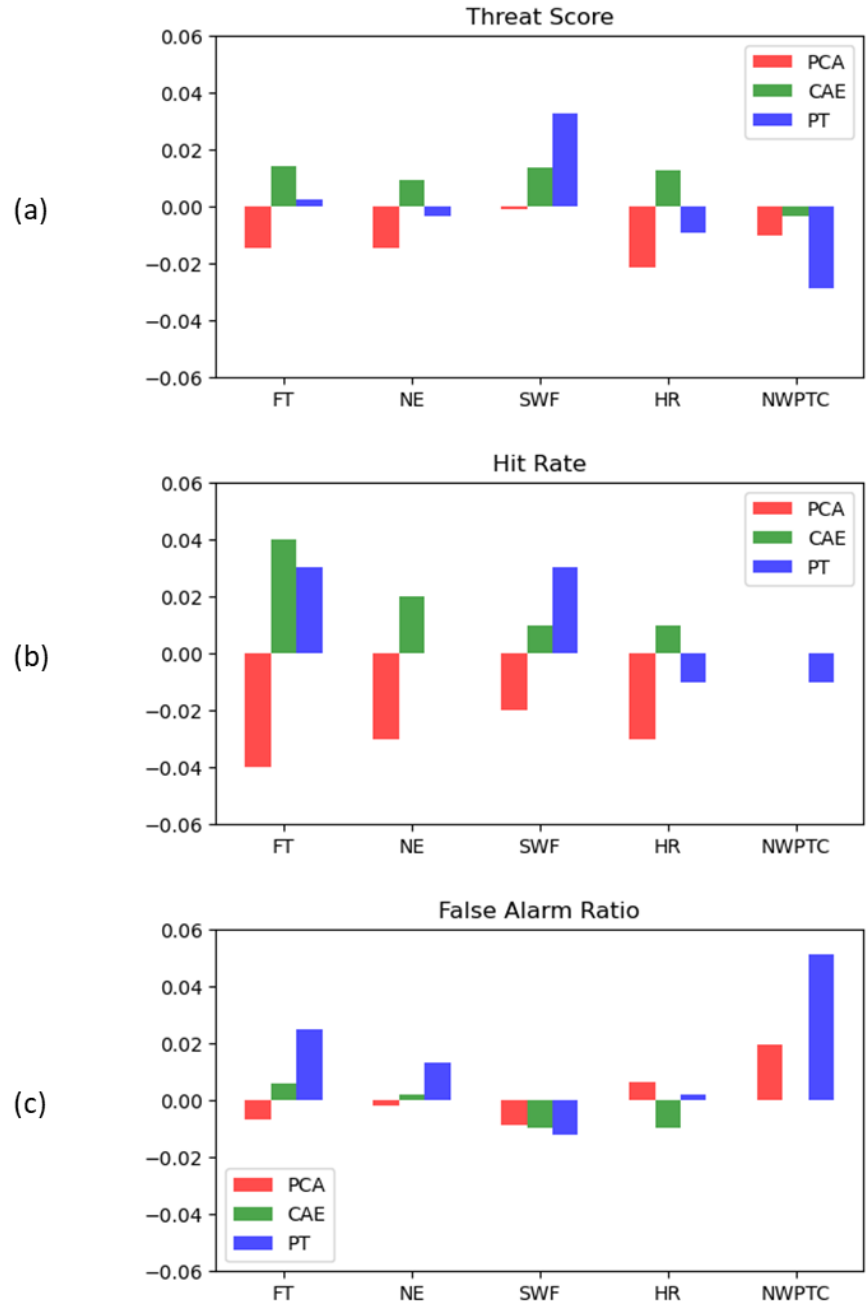

Figure 4. The change in evaluation metrics between high-resolution (512x512) and low-resolution (256x256) datasets. A positive value means the evaluation metric of the high-resolution experiment is higher than that in the low-resolution configuration. Positive values of threat score (panel a) and hit rate (panel b) and negative values of false-alarm rate (panel c) represent an improvement while using high-resolution data.

4.3 Experiment 3: the sizes of the latent space

In experiments 1 and 2, we forced the dimension of latent spaces to be 2048. This number is set to be consistent with the pre-trained model (PT). For algorithms other than pre-trained models, will the performance be different if we change the sizes of the latent spaces?

We conducted the same classification tasks for latent space dimensions ranging from 4 ($2^2$) to 2048 ($2^{11}$), and the results are shown in Figure 5. Figure 5 indicates that the hit rate (POD) didn't change much when using smaller latent space. However, the false-alarm rate increased, so the threat score dropped. This trend is consistent for both methods, while the optimal latent space size differs for various weather events.

Another observation from Figure 5 is that PCA seemed more robust than CAE when using a smaller latent space dimension. Take the FT event, for example; if we look at the threat score (the black lines), CAE (the dashed line) underperformed PCA (the solid line) when the dimension size was smaller than 128. This crossover varied in other events, but CAE always lost advantages when the latent space dimension was small. Moreover, the change in the classification metrics for PCA is much smoother than for CAE, which indicates the classic linear transformation algorithm is more robust in nature.

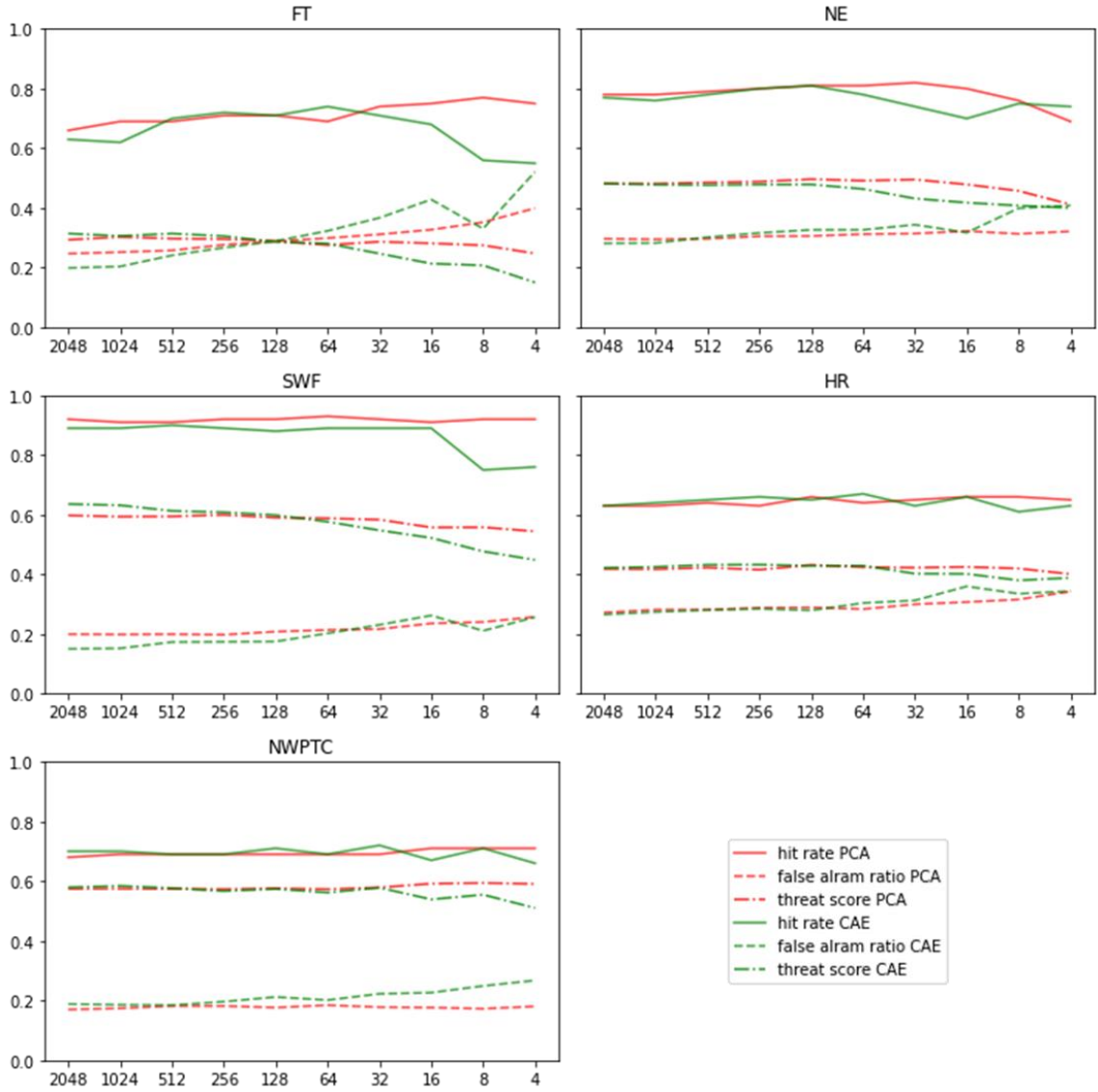

Figure 5. The threat scores (dash-dotted line), the hit rates (solid line), and the false-alarm rates (dashed line) of the classification with different sizes of the latent space. The PCA is colored in red, and CAE in green.

## 5 Discussion and Conclusions

The experimental results shown in the previous section met our expectations. As shown in the results, CAE consistently outperformed other algorithms in different experimental configurations. Furthermore, the results suggested higher resolution images improve the classification performance, especially for CAE. We also tested the effect of the sizes of the latent space and found that using a smaller latent space is feasible for the designed tasks. In this section, we will further look into the reconstruction, interpretability, and computational cost of the investigated algorithms.

5.1 The reconstruction from the latent space

Among the investigated representation learning methods, both PCA and CAE provide mechanisms to reconstruct the data from the latent space. We selected one case for each weather type and illustrated the original data (left panel), the reconstruction with the first 2048 principal components (the center panel), and that with the CAE (the right panel) in Figure 5. Figure 5 shows that both reconstruction methods keep the general pattern and lose fine details, which is expected since we compress the data size from 65,535 points to 2,048. However, while the reconstruction with CAE represented a smooth and blurry version of the original image, the PCA reconstruction exhibited high-frequency noises in the figures. Such results are expected when applying PCA to spatial data because the low-frequency modes usually come with larger eigenvalues; hence, our reconstructions remove certain high-frequency information (Novembre and Stephens, 2008).

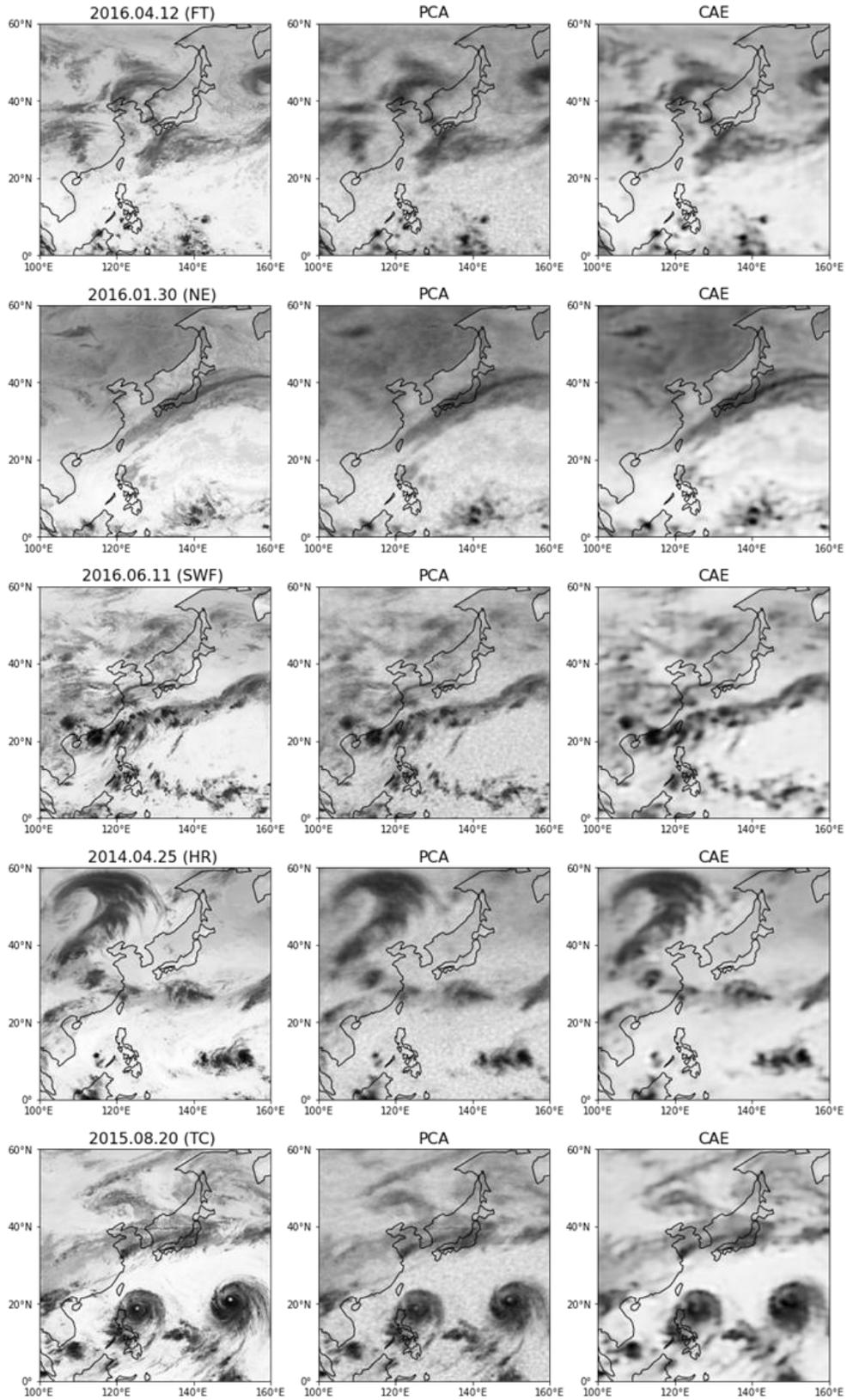

Figure 6. The original GRidSat-B1 images (left column) and their reconstructions (center column for PCA and right column for CAE) of five selected cases.

5.2 The interpretation of the representations

The approach proposed in this study combined learned representations and a generalized linear model to identify weather events. The significance tests of GLM can indicate the importance of the learned features. Therefore, for each classification task, the proposed framework can lead to an interpretable model as long as we can interpret the learned representations. For example, table 2 summarizes the GLM results of using CAE-derived feature vectors to predict the SWF event. Here we use the latent space size of 8 for readability. In table 2, we see that feature 1, 2, and 5 pass the significance test at the level of $P < 0.001$ and may be worth further investigation. The same analysis can be performed with feature vectors derived from PCA and other representation-learning algorithms.

Although we used GLM in this study, other classification algorithms that can indicate the relative importance of predicting variables, e.g., tree-based algorithms such as random forest (Breiman, 2001) and gradient boosting machine (Friedman, 2001), can also serve as alternatives.

As discussed in the method section, PCA has long been used in atmospheric science studies. Each principal component can be directly illustrated on the map and interpreted by domain experts. In contrast, autoencoders mapped the data into an abstract latent space where each dimension is a nonlinear mapping of the input space. Thus, direct visualization of the axis of the latent vectors may not be human-readable, hindering its interpretability.

Table 2. The summary table of the generalized linear model (logistic regression) for predicting the SWF event using CAE-derived features.

| Feature | Coeficient | Std-Error | z | P>|z| | 95% CI |
|---|---|---|---|---|---|
| 0 | 0 | 0 | NA | NA | [0, 0] |
| 1 | 4.34e-1 | 5.1e-2 | 8.489 | 0.000 | [0.334, 0.535] |
| 2 | 8.96e-17 | 8.6e-18 | 10.425 | 0.000 | [7.4e-17,1.2e-16] |
| 3 | 4.16e-2 | 7.0e-2 | 0.596 | 0.551 | [-0.095, 0.178] |
| 4 | -1.04e-17 | 8.6e-18 | -1.199 | 0.230 | [-2.7e-17, 6.57e-18] |
| 5 | -3.99e-1 | 3.0e-2 | -13.206 | 0.000 | [-0.458, -0.339] |
| 6 | 0 | 0 | NA | NA | [0, 0] |
| 7 | 7.6e-3 | 3.0e-2 | 0.255 | 0.799 | [-0.051, 0.066] |

5.3 The computational costs

Finally, we want to note the investigated algorithms' computational cost as a reference. All experiments of this study were conducted on a server with 12 CPU cores of 3.7GHz. The server has 64GB memory and an NVIDIA RTX-2080Ti GPU for accelerating deep neural network computation. The computational and storage costs are summarized in Table 3. As shown in Table 3, CAE is the most affordable method for computation time and storage space, given that GPU acceleration is available. The software package used for PCA by default uses all CPU cores and gets decent acceleration. We also conducted another deep learning method in the original experimental design, Variational Autoencoder (CVAE). However, the classification results were not comparable to other methods and hence were not shown in the report.

Table 3. The computational cost of PCA and CAE.

|  |  | PCA | CAE |
|---|---|---|---|
| Learning Time | (256x256) | ~26 minutes | ~6 minutes |
|  | (512x512) | ~183 minutes | ~23 minutes |
| Storage | (256x256) | 1.1GB | 1.8MB |
|  | (512x512) | 4.1GB | 6.2MB |
| CPU |  | 12 | 1 |
| GPU acceleration |  | No | Yes |

5.4 Concluding Remarks

In this study, we investigated representation learning algorithms on satellite images and evaluated the learned latent spaces with classifications of various weather events. The experiment results suggested that the convolutional autoencoder (CAE) can effectively project the data into latent spaces and showed the highest threat scores in all tasks. At the same time, the classic linear transform, PCA, yielded a similar hit rate but a higher false-alarm rate. The pre-trained model performed exceptionally well at recognizing tropical cyclones but was inferior in other tasks.

The classification performance for different weather events varied depending on how relevant their definitions are to the brightness temperature. For example, while SWF events and tropical cyclones usually occur with significantly high clouds, their hit rates and threat scores are much higher than subjectively defined events such as front.

Further experiments suggested that representations learned from higher-resolution datasets are superior in all classification tasks, and the CAE can benefit more than other algorithms. We also found that smaller latent space sizes had little impact on the classification task's hit rate as long as the dimension size was larger than 128. However, a small latent space dimension could cause a significantly higher false-alarm rate.

In terms of interpretability, the features learned by PCA can be easily visualized in the physical domain and interpreted by domain experts. In contrast, though the visualization of CAE is possible, the lack of a direct connection to physical attributions could be the weakness of this approach.

The convolutional autoencoder (CAE) is an effective and efficient representation learning algorithm. The feature vectors learned with CAE showed good performance in various classification tasks, and its performance benefits from high-resolution satellite images more than other algorithms. However, its lack of physical interpretability suggested further studies on incorporating physics terms into the deep neural network algorithms to construct efficient and physically interpretable representations.

Finally, we want to comment on the implications of our work for disaster reduction. While a high hit rate in identifying extreme weather events is crucial, our results suggested that both PCA and CAE with a small latent space size can be useful for risk management. If we consider the future availability of high-resolution and multiple-modal data, CAE is a technology worth investing in.